\documentclass[letterpaper, 10 pt, conference]{ieeeconf}  

\IEEEoverridecommandlockouts                              

\newcommand*\diff{\mathop{}\!\mathrm{d}}
\usepackage[utf8]{inputenc}
\usepackage[T1]{fontenc}
\usepackage{amsmath,color}
\usepackage{amsfonts}
\usepackage{multirow}
\usepackage{algorithm}
\usepackage{algpseudocode}
\usepackage{xifthen}
\newcommand{\comment}[2][]{{\color{red}
		\ifthenelse{\isempty{#1}}%
		{}
		{\textit{#1}:~}
		#2}}

\usepackage{tabularx}
\newcolumntype{L}[1]{>{\hsize=#1\hsize\raggedright\arraybackslash}X}%
\newcolumntype{R}[1]{>{\hsize=#1\hsize\raggedleft\arraybackslash}X}%
\newcolumntype{C}[1]{>{\hsize=#1\hsize\centering\arraybackslash}X}%
\usepackage{xcolor}
\usepackage[colorlinks]{hyperref}
\usepackage[switch, modulo]{lineno}

\newcommand{\arcangle}{%
	\mathord{<\mspace{-9mu}\mathrel{)}\mspace{2mu}}%
}

\usepackage[pdftex]{graphicx}
\graphicspath{{./}{./images/}{./figures/}}

\begin{document}

\title{\LARGE \bf
	Ergodic exploration of dynamic distribution}

\author{Luka~Lanča, Karlo~Jakac, Sylvain~Calinon and Stefan~Ivić*
	\thanks{L.~Lanča, K.~Jakac and S.~Ivić are with Faculty of Engineering, University of Rijeka, Rijeka, Croatia.}
	\thanks{S. Calinon is with Idiap Research Institute, Martigny, Switzerland.}
	\thanks{*Corresponding author, e-mail: {stefan.ivic@uniri.hr}.}
}

\maketitle
\thispagestyle{empty}
\pagestyle{empty}

\begin{abstract}
This research addresses the challenge of performing search missions in dynamic environments, particularly for drifting targets whose movement is dictated by a flow field. This is accomplished through a dynamical system that integrates two partial differential equations: one governing the dynamics and uncertainty of the probability distribution, and the other regulating the potential field for ergodic multi-agent search. The target probability field evolves in response to the target dynamics imposed by the environment and accomplished sensing efforts, while being explored by multiple robot agents guided by the potential field gradient. The proposed methodology was tested on two simulated search scenarios, one of which features a synthetically generated domain and showcases better performance when compared to the baseline method with static target probability over a range of agent to flow field velocity ratios. The second search scenario represents a realistic sea search and rescue mission where the search start is delayed, the search is performed in multiple robot flight missions, and the procedure for target drift uncertainty compensation is demonstrated. Furthermore, the proposed method provides an accurate survey completion metric, based on the known detection/sensing parameters, that correlates with the actual number of targets found independently.
\end{abstract}


\section{Introduction}

Despite significant advancements in maritime technology, accidents at the sea continue to occur, highlighting the ongoing need for search and rescue (SAR) missions. The vast inspection area and the dynamic nature of ocean movement, introducing additional complexity and uncertainty, make search missions exceptionally difficult \cite{futch2019search}. A critical factor when considering maritime SAR missions is time, since the survival rate of missing individuals decreases over time. 

To address some of these challenges, unmanned aerial vehicles (UAVs) have proven to be useful in SAR operations, particularly in maritime environments, as demonstrated in \cite{lun2022target, cho2021coverage, yang2020maritime}. Various approaches have been proposed, focusing on different aspects such as probability field generation, dynamic target tracking, multi-agent coordination, and search strategies. An algorithm for intelligent maritime response plan generation is proposed in \cite{ai2019intelligent}, some of which are analyzed on real maritime SAR scenarios. Regarding camera sensing systems, a review of methods for automatic target detection in maritime SAR aerial images is presented in \cite{martinez2025maritime}. A method for oceanic search using dynamically changing target probability maps, relying on Gaussian mixture models, is presented in \cite{lun2022target}. The method also incorporates a time-varying ocean velocity field, but they utilize grid-based domain decomposition. Similarly, \cite{li2021cooperative} applies a grid-based method for dynamic target search using multiple UAVs. It considers UAV communication flaws and data loss. Related to communication, in \cite{yang2020maritime}, UAVs and Unmanned Surface Vehicles (USVs) are used to construct a temporary communication network when performing maritime SAR missions. It also utilizes a grid method and generates a path plan based on information sea maps and sensor data. Multi-agent ergodic search utilizing dynamic information maps is presented in \cite{coffin2022multi}. The search agents utilize low information sensors for localization and tracking of moving targets. Genetic algorithm optimization for locating moving targets with UAVs is utilized in \cite{alanezi2022dynamic}. The search area is defined as a grid, and the belief map is updated using a Bayesian approach based on sensing effects. Another technique that utilized grid-based domain decomposition is presented in \cite{zhang2023usv}. It utilizes USVs for performing lawnmower pattern search, but it lacks dynamic probability field evolution. The work in \cite{cho2021coverage} provides real world experimental results of sea search missions where the UAVs' path is pre-computed and then forwarded for execution. It does not feature probabilistic sensing or dynamic probability field. 

A logical approach to conducting SAR missions is to prioritize searching areas with higher probability of locating the target, which can be achieved by employing ergodic search methods. By specifying a metric of the search success, and its maximization during the ergodic search process, a specific coverage of the target probability distribution can be ensured, effectively distinguishing between explored regions and those that require further exploration. Most commonly used ergodic search methods are Spectral Multiscale Coverage (SMC), Model Predictive Control (MPC), and Heat Equation Driven Area Coverage (HEDAC). SMC, proposed in \cite{mathew2011metrics}, utilizes smoothed Fourier basis functions in order to generate ergodic trajectories for multiple search agents. The method has been adapted for conducting missions in dynamical environments in \cite{mathew2010uniform}, and utilized in \cite{ivic2020search}, where the authors proposed a search strategy for the missing MH370 plane that crashed in 2014 in the Indian Ocean. The dynamics of the probability density field were approximated using moving Lagrangian particles. Another commonly used approach is MPC, which formulates the generation of ergodic trajectories as an optimization problem \cite{abraham2018decentralized}. In \cite{abraham2017ergodic}, it was utilized for area exploration and objects' shape estimation. An example of MPC ergodic control for searching dynamically changing information distribution is presented in \cite{mavrommati2017real}, where it is applied to a coverage and localization problem with targets' motion modeled as a diffusion process. Beyond that, another method which can be used is HEDAC, which was firstly introduced in \cite{ivic2016ergodicity}. It was additionally formulated with the Finite Element Method (FEM) which allowed its use in irregularly shaped domains with support for inter-domain obstacles \cite{ivic2022constrained}, and expanded with probabilistic sensing methods \cite{ivic2020motion}. HEDAC is also employed in real-world robotics applications, such as cleaning curved surfaces with a robotic manipulator \cite{Bilaloglu25TRO}. To the authors knowledge, the method has not yet been utilized for searching targets that exhibit dynamic behavior.  

The goal of this letter is to improve the methodology for dynamic search, particularly for tracking drifting targets in complex ocean currents. It effectively integrates an evolving Eulerian probability density model with a potential-based ergodic search control. The method is designed for use with robot agents, such as Unmanned Aerial Vehicles (UAVs), that are not affected by the ocean flow. Additionally, the diffusion term allows for compensation of uncertainty caused by potential inaccuracies in the measured flow field. Ultimately, the proposed methodology enables real-time estimation of the total target detection probability, which is crucial for managing and making decisions in ocean SAR missions.

\section{Dynamic probability distribution field and sensing}

Let us consider a multi-agent exploration of the domain $\Omega \in \mathbb{R}^2$ in time $t$. For now, the agents' trajectories and directions, denoted by $\mathbf{z}_i(t)$ and $\theta_i(t)$ respectively, are considered known, where $i = 1, \dots, n$ is the index of each agent, and $n$ is the total number of agents conducting exploration.

The search is conducted based on the probability density of undetected target presence $m(\mathbf{x},t): (\Omega \times t) \to \mathbb{R}$. Its initial value, at $t=0$, is denoted as $m_0(\mathbf{x})$, and satisfies the condition
\begin{equation}
	\int_\Omega m_0 \diff\mathbf{x}=1. 
	\label{eq:m_normalization}
\end{equation}

The environment dynamics are defined with the field $\mathbf{w}(\mathbf{x}, t): (\Omega \times t) \to \mathbb{R}^2$, representing fluid flow that causes the drifting effects of the targets. Both $m_0(\mathbf{x})$ and $\mathbf{w}(\mathbf{x}, t)$ are assumed to be known.

Each agent actively explores the domain and its sensing effect is modeled with a sensing function $\gamma_i(\mathbf{r})$, where $\mathbf{r}(t)$ are coordinates in the agent's local coordinate system. The sum of all sensing applied by the agents in the global coordinate system is represented with
\begin{equation*}
	\Gamma(\mathbf{x}, t) = \sum_i^n  \gamma_i\left(\mathbf{R}(\theta_i(t))\cdot\left(\mathbf{z}_i(t) - \mathbf{x}\right)  \right),
	\label{eq:coverage_b}
\end{equation*}
where $\mathbf{R}$ is the rotation matrix defined as
\begin{equation*}
	\mathbf{R}(\theta)={\begin{bmatrix}\cos \theta &-\sin \theta \\\sin \theta &\cos \theta \\\end{bmatrix}}.
\end{equation*}

The movement of the search targets is governed by the vector field $\mathbf{w}(\mathbf{x},t)$, which imposes dynamic behavior of $m$. To effectively simulate $m$ and its uncertainties, we introduce an advection-diffusion partial differential equation including a sink term simulating the sensing effects of search agents. The dynamics of the probability of undetected target presence $m(\mathbf{x},t)$ can be defined as
\begin{equation}
	\frac{\partial m}{\partial t} = D\cdot \nabla^2 m - \mathbf{w}\cdot\nabla m - \Gamma\cdot m,
	\label{eq:pde_m}
\end{equation}
where $D$ is the diffusion coefficient. Diffusion effectively models the collective uncertainty of the system which can result from errors of the underlying advection flow field, errors in setting up the initial undetected target probability field, and errors in agent localization which directly effects its sensing. The diffusion coefficient is calculated by using the equation for mean square displacement of a brownian particle, represented with 
\begin{equation*}
	E^2 (t) = 2 \cdot D \cdot t,
\end{equation*}
where $E$ represents the mean displacement of a particle for time $t$. To better align this with a physical representation of uncertainty, the diffusion coefficient will be calculated based on the approximated distance error of targets drift $e$, for time $t$. It can also be determined with the measured flow field error value in m/s, if that estimate is known. The diffusion coefficient for the two-dimensional case can now be expressed as
\begin{equation}
	D = \frac{e^2}{2 \cdot t}.
	\label{eq:diff_coef}
\end{equation}

The exploration success is evaluated with the survey accomplishment metric 
\begin{equation*}
	\eta(t) = 1 - \int_\Omega m(\mathbf{x}, t) \diff \mathbf{x}.
\end{equation*} 
Since the obvious goal of the search is a rapid and continuous minimization of survey accomplishment $\eta$, we can suitably define an ergodic search task as
\begin{equation*}
	\lim_{t\to \infty} \int_\Omega m \diff\mathbf{x} = 0.
\end{equation*}

Note that the search cannot continue indefinitely, however, its duration cannot be defined in advance. Since the duration is unknown, the dynamic distribution exploration cannot be set up as an optimization problem, and instead we consider it as an ergodic task.

\section{Motion model}

The agents' motion is modeled with the Dubins model by appointing constant velocity $v_i$ to each agent and varying their heading angle $\theta_i$. Therefore, the agents' trajectories can be represented with 

\begin{equation}
	\begin{aligned}
		\frac{\diff\mathbf{z}_i}{ \diff t} & =
		\left[
		\begin{array}{c}
			v_{i} \cdot \cos\theta_i \\
			v_{i} \cdot \sin\theta_i
		\end{array}
		\right].
	\end{aligned}
	\label{eq:motion_model}
\end{equation} 

The change of the heading angle is governed with the yaw angular velocity 

\begin{equation*}
	\omega_i(t) = \frac{\diff \theta_i}{\diff t},
\end{equation*}
constrained by $\left|\omega_{i}\right| \leq \omega_{i}^{max}$, which also determines the agents' minimal turning radius $r_{i}^{min} = \frac{v_i}{\omega_{i}^{max}}$. 

It's important to note that we consider a search with flying robot agents, so the flow filed $\mathbf{w}$ has no effect on the agents' motion. 

The motion control is governed by the HEDAC algorithm \cite{ivic2022constrained}, which calculates $\omega(t)$ based on the potential field $u(\mathbf{x}, t)$ obtained by solving the partial differential equation 
\begin{equation}
	\alpha \nabla^2 u (\mathbf{x}, t) -u (\mathbf{x}, t) + m (\mathbf{x}, t) = 0,
	\label{eq:hedac_pde}
\end{equation} 
where $\alpha>0$ is a parameter which regulates the balance between global and local exploration. On the domain boundary $B$ (and possible obstacles) Neumann boundary conditions are set as
\begin{equation*}
	\left.\frac{\partial u}{\partial\mathbf{n}}\right|_B = 0,
	\label{eq:hedac_bc}
\end{equation*}
where $\mathbf{n}$ is outward normal vector to boundary $B$.

For directing the exploration agents we suitably calculate the unitary gradient of the potential $u$:

\begin{equation*}
	\mathbf{u}(\mathbf{x}) = \frac{\nabla u(\mathbf{x})}{||\nabla u(\mathbf{x})||}.
\end{equation*}

By utilizing the gradient of the potential field $\mathbf{u}$, the yaw angular velocities can be computed with

\begin{equation}
	\omega_{i} = \frac{\diff}{\diff t} \left(\arcangle \left(\theta_i, \;\mathbf{u}(\mathbf{z}_i) \right)\right).
	\label{eq:dubins_angular_velocity}
\end{equation}
The angular velocity $\omega_{i}$ can be either positive or negative, depending on the turning direction: a positive value indicates counterclockwise rotation, while a negative value represents a clockwise rotation.

The last step is performing obstacle avoidance procedure, which is described in detail in \cite{ivic2022constrained}, to approve or modify $\omega_{i}$ that results in the final yaw angular velocity value, which guarantees a collision free trajectory.

\section{Implementation}
\label{sec:implementation}

The implementation of the proposed methodology is achieved by coupling the solvers for the probability density advection/diffusion/sensing in \eqref{eq:pde_m}, the potential field governed by \eqref{eq:hedac_pde}, and search agents' motion model \eqref{eq:motion_model}. The advection and diffusion terms in \eqref{eq:pde_m} are solved using Finite Volume Method (FVM), with the \textit{scalarTransportFoam} solver included in \textit{OpenFOAM}. 
Meanwhile, sensing is explicitly applied to the field $m$ based on agents positions $\mathbf{z}_i$ and their sensing functions $\gamma_i$ in each time step. FEM-based solver \textit{Netgen/NGSolve} \cite{schoberl1997netgen, schoberl2014c++} is utilized for solving partial differential equation \eqref{eq:hedac_pde}, and obtaining the gradient of the potential field. Solving the motion model, collision avoidance procedure, different utility and visualization operations, and integrating all components is done using the Python programming language. 

In method's implementation, motion control and sensing are applied with a period of $\Delta t$, while the advection/diffusion simulations are performed with a smaller time step of $\Delta t / 10$ for a duration of a control step $\Delta t$. 

The two coupled methods inherently work differently, HEDAC uses FEM, while \textit{OpenFOAM} uses FVM. In order to adapt them to the problems they solve, the underlying meshes are differently structured. The HEDAC FEM utilizes two-dimensional triangular mesh, while \textit{OpenFOAM} uses a three-dimensional hexahedral mesh (with a single element in the third dimension) for solving a two-dimensional advection/diffusion case. In addition, FEM stores values at the nodes, whereas the FVM stores them at the cell centers, requiring data value transfer. The methods are combined in a way that the agents' sensing effects are directly applied to the FVM scalar field, and then the $m$ field is projected to the FEM mesh for the potential field calculation. The projection is executed by using nearest neighbor interpolation method due to its computational efficiency. Only probability field $m$ accumulates for an error, given that the potential field $u$ is recalculated from updated values $m$ in each time step. Hence, we consider the use of nearest neighbor method and its resulting coarseness to be justified. 

The entire procedure for the proposed methodology, including all vital subprocedures and theoretical references, is provided in detail in Algorithm \ref{alg:m}.

The advection-diffusion framework employs \textit{OpenFOAM-v2406} \cite{openfoam}, with case meshes created using either the \textit{blockMesh} utility or \textit{cfMesh}. The HEDAC FEM framework was implemented using \textit{Python 3.12}, \textit{NumPy 2.1}, \textit{NGSolve 6.2}, \textit{Netgen}, \textit{SciPy}, and \textit{Gmsh}. The visuals are generated using \textit{Matplotlib}, \textit{PyVista}, \textit{CMasher}, and \textit{Seaborn}. All the simulations were computed on a notebook PC with 6-core 2.60 GHz CPU, 16 GB of RAM and SSD hard drive.  

\begin{algorithm}
	\footnotesize
	\caption{Procedure for ergodic exploration of dynamic distribution}
	\label{alg:m}
	\begin{algorithmic}	
		\Procedure{Motion control with dynamic probability field}{}      
		
		\Function{initialization}{}
		\State Initialize general parameters, agents, FEM 
		\State Normalization and setting $m$ field in FES \Comment{\eqref{eq:m_normalization}}
		\State Saving normalized $m$ field to the the OpenFOAM case for $t=0$
		\State Setting OpenFOAM vector field $\mathbf{w}$
		\State Setting OpenFOAM case diffusion coefficient
		\State $t=0$
		\EndFunction
		
		\Function{Solve trajectories}{}
		\State Obtain scalar field $m(t)$ from OpenFOAM case
		\For{$i = 1 \ldots n$} \Comment{For all agents}
		\State Apply sensing to field $m(t)$ \Comment{sensing in \eqref{eq:pde_m}, $\gamma_i$} 
		\EndFor
		\State Saving $m$ field with applied sensing to the OpenFOAM case
		\State Updating $m$ field in FES
		\State Solving potential $u$ \Comment{Equation \ref{eq:hedac_pde}}
		\State Calculating agent's directions \Comment{\eqref{eq:dubins_angular_velocity}}
		\State Collision avoidance procedure \Comment{Equivalent to \cite{ivic2022constrained}}
		\State Updating agent's positions
		\State Scalar transport of $m$ in OpenFOAM for $\Delta t$ \Comment{adv./diff. in \eqref{eq:pde_m}}
		\State Increase the time: $t = t + \Delta t$
		\EndFunction
		
		\EndProcedure		
	\end{algorithmic}
\end{algorithm}

\section{Test cases}

The method was tested in simulations based on two search scenarios. The first scenario showcases a search mission in a rectangular domain where the dynamics of the targets is imposed by a cavity lid-driven flow. Furthermore, it demonstrates search performance gains by using the proposed dynamic target probability density field compared to the static field. The second scenario represents a simulation of a realistic ocean search scenario using UAVs. The targets' dynamic behavior is governed by the transient ocean flow field. The search start is delayed to account for the movement of targets between their last recorded location and the actual start of the search. Additionally, the search is completed in multiple phases, taking into account the time needed for UAV battery replacement, and the targets motion during that time. The second test case also showcases the benefits of diffusion utilization for managing uncertainty. 

Supplementary materials and video animations for the computed test cases are available on the Open Science Framework repository: \url{https://osf.io/3a6xc/}.

\subsection{Search simulations setup and evaluation metrics}

Each case contains a total of 1000 simulated targets which are advected with the flow field $\mathbf{w}$ and the initial distribution of targets at $t=0$ matches the initial probability of undetected target presence $m_0$. The search targets' positions remain consistent over time across all simulations that share the same flow field, and it is governed by particle Lagrangian motion law
\begin{equation*}
	\frac{\diff\mathbf{y}}{\diff t} = \mathbf{w}(\mathbf{y}, t).
	\label{eq:lagrangian_motion}
\end{equation*}
Search scenarios are evaluated using two metrics: the survey accomplishment metric $\eta$, which reflects the system's perception of survey completion, and the target detection rate $\kappa$, which represents the proportion of detected targets relative to the total number of targets, providing a realistic measure of survey completion.

\subsection{Synthetic case - Cavity flow}

\begin{figure}[!ht]
	\centering
	\includegraphics[width=0.7\linewidth]{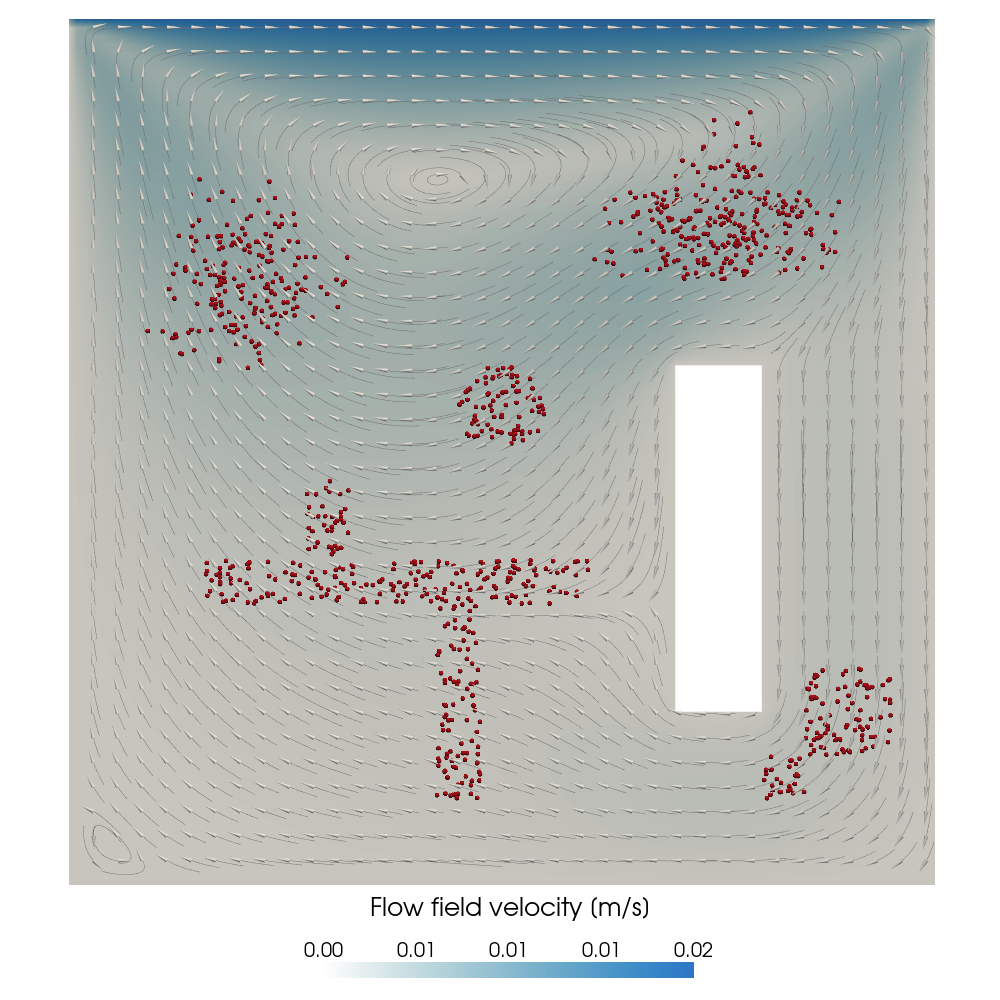}
	\caption{Cavity lid-driven flow field represented with a vector field, streamlines, and velocity intensity contour plot. Positions of simulated search targets at $t=0$ are displayed as red dots. }
	\label{fig:cavity_flow}
\end{figure}

The search domain is defined with an $1\times1$ m square which includes a rectangular obstacle represented with a bounding box from (0.7, 0.2) m to (0.8, 0.6) m. The flow field inside the domain is considered as known. It was computed with the \textit{simpleFoam} solver utilizing the \emph{k-$\omega$ SST} turbulence model. The velocity boundary condition on all rectangle sides was set to \textit{no slip} condition, implying that the fluid that is touching the boundary has a velocity of 0 m/s in reference to that boundary, and the pressure boundary condition is set as \textit{zero gradient}. The fluid motion is caused by moving the upper boundary of the domain rectangle at the velocity of 2e-2 m/s, and the kinematic viscosity of the fluid  which is equal to 1e-6 $\text{m}^2/\text{s}$. The flow field, represented with a vector field, matching streamlines, and contour plot of velocity intensity is displayed in Fig. \ref{fig:cavity_flow}. It is steady throughout the whole duration of the search, and the average flow field velocity magnitude is 3e-4 m/s.

The search targets are randomly distributed across five distinct areas of varying shapes. In the two upper areas, the targets follow a standard distribution centered within each area, whereas in the remaining areas, they are uniformly distributed. The positions of the simulated targets are shown in Fig. \ref{fig:cavity_flow}.

The domain is explored with 3 identical search agents for the time $T = 900$ s, with the search and agents' motion parameters provided in Table \ref{tab:agent_charactristics}. 

Each agent is equipped with a sensor that has a circular detection radius $r_d=0.015$ m. The target is detected in one flyover with a chance of $\mu_a = 0.8$, if the target is directly below the search agent, but the detection chance drops from that point following the standard distribution with a standard deviation of $\sigma_a = 0.015$ m. The corresponding sensing rate function in the local (agent's) coordinate system can be described with

\begin{equation*}
	\gamma_a = -\frac{\ln(1 - \mu_a)}{2} \cdot \mu_a \cdot e^{-0.5 \left( \frac{d}{\sigma_a} \right)^2},
\end{equation*}
where $d$ is the distance from the current agent's location $\mathbf{z}_i(t)$ and the observed point. 

\begin{table}[h!]
	\scriptsize
	\centering
	\caption{Control and motion parameters used in simulated search scenarios}
	\begin{tabularx}{\linewidth}{L{2.95}R{0.6}R{0.6}R{0.6}L{0.25}} 
		Search  parameters									& Cavity 		& Unije Channel			& Units				\\
		\hline 	
		Agents' velocity $v$							& 0.015					& 10			& m/s					\\
		Agents' minimum turning radius $R_{min}$			& 0.01					& 100			& m					\\
		Agents' minimum clearance distance $\delta$			& 0.01					& 50			& m					\\
		
		Time step 	$\Delta t$								& 0.2					& 3				& s					\\
		Alpha $\alpha$ 										& 5e-2					& 1e5			& -					\\

	\end{tabularx}
	\label{tab:agent_charactristics}
\end{table}

\begin{figure*}[!h]
	\centering
	\includegraphics[width=0.92\linewidth]{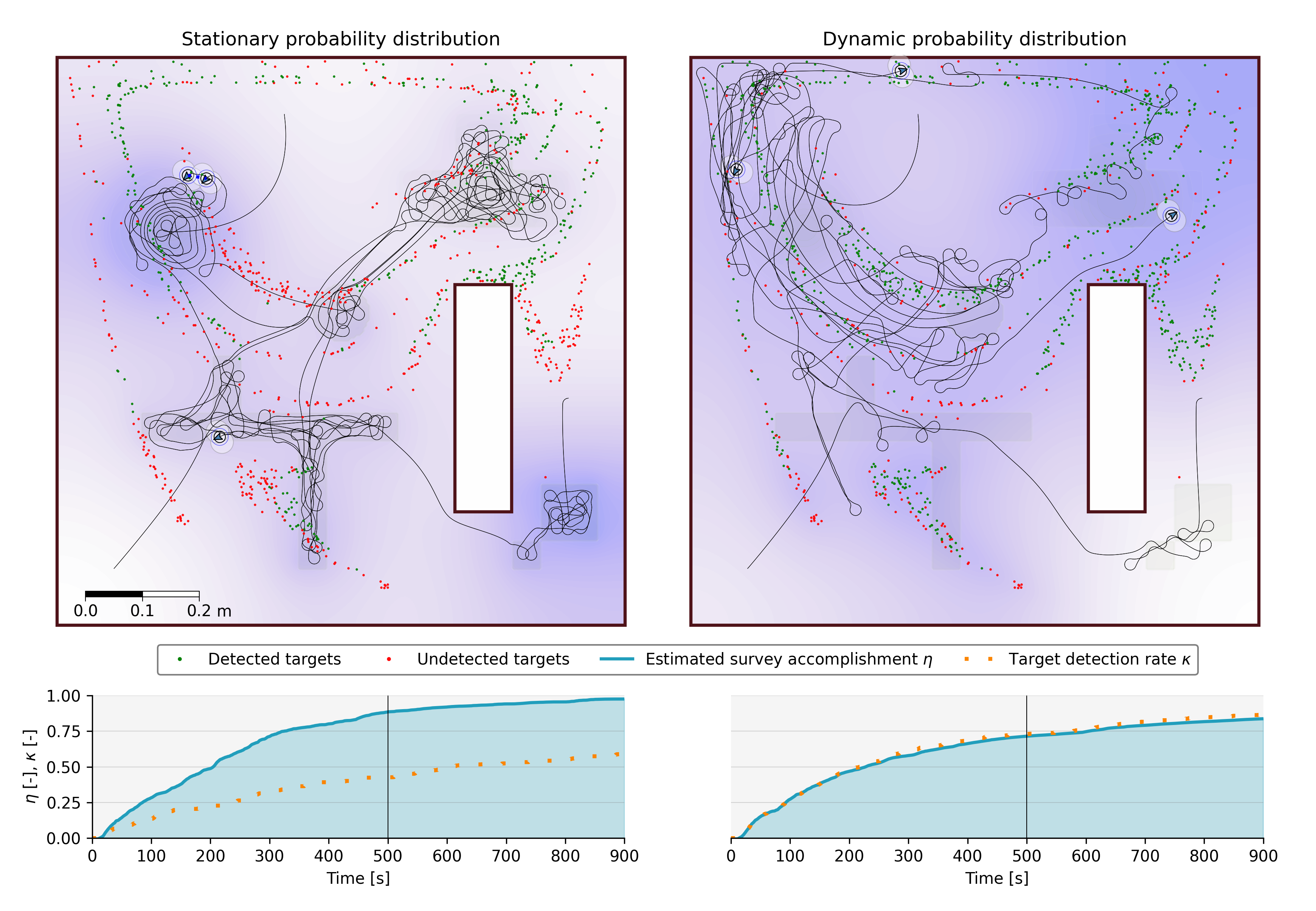}
	\caption{Comparison of the static probability search and the proposed method. The figure features the search domain at $t=500$ s, with the underlay of the potential field displayed with the purple-white shaded heat-map for both cases. It displays agents' trajectories and the simulated search targets. Under the domain plots, the figure includes a search performance analysis graphs containing survey accomplishment $\eta$ and target detection rate $\kappa$ metrics in relation to search time. The difference in the agent behavior recognized by observing the agents' trajectories, where the agents in the static probability scenario focus their search only on the area covered by the initial target probability distribution $m_0$, while in the agents' motion in the dynamic case. The search performance improvement can be seen by comparing the target detection rate between the cases, furthermore, the survey accomplishment in the proposed method stays true to the target detection rate.}
	\label{fig:cavity_showcase_frame}
\end{figure*}

The search is performed using the proposed method, which considers the dynamic properties of the probability distribution influenced by the underlying flow field. This approach is compared to a baseline search conducted with a method assuming a stationary target probability distribution. The results of both simulations are compared and shown in Fig. \ref{fig:cavity_showcase_frame}. Predictably, the proposed method demonstrated superior performance, achieving a target detection rate $\kappa$ approximately 50\% higher than the baseline throughout the entire duration of the search simulation. The survey accomplishment metric $\eta$ corresponds to $\kappa$ in the simulation conducted with the proposed method, reflecting an accurate perception of survey completion and representing the realistic state of the search. In contrast, for the baseline method, $\eta$ does not align with $\kappa$, as the algorithm indicates much higher survey completion than is realistically achieved.

Additionally, an analysis was conducted to examine the impact of the flow field's and search agents' velocities on the search performance in both scenarios, using the proposed and baseline methods. To conduct this analysis, a new variable, $\lambda$, is introduced, representing the ratio of the agents' velocity to the flow field's average velocity. This variable highlights the relationship between the agents' velocity and the flow field.

The analysis was completed using the same cavity case since, conveniently, the target probability and the targets circulate inside the domain and are unable to escape it. The survey performance was analyzed for different values of $\lambda$ in range $[0.25, 1000]$. Lower $\lambda$ values indicate that the flow field's average velocity is higher relative to the agents' velocities, and vice versa. To achieve achieve different $\lambda$ values, the flow field used in the test case presented above was scaled while the agent velocity remained constant. This ensured that only the intensity of the flow field was modified, while its structure remained unchanged.

For reference, the realistic $\lambda$ value for performing a search mission in the sea coastal region would be somewhere around the value of 50, considering a search with multi-rotor UAVs with a  velocity of 10 m/s searching in a submesoscale  sea flow with an average velocity of 0.2 m/s \cite{jakac2025efficient, cosoli2013surface}.

\begin{figure}[!ht]
	\centering
	\includegraphics[width=0.85\linewidth]{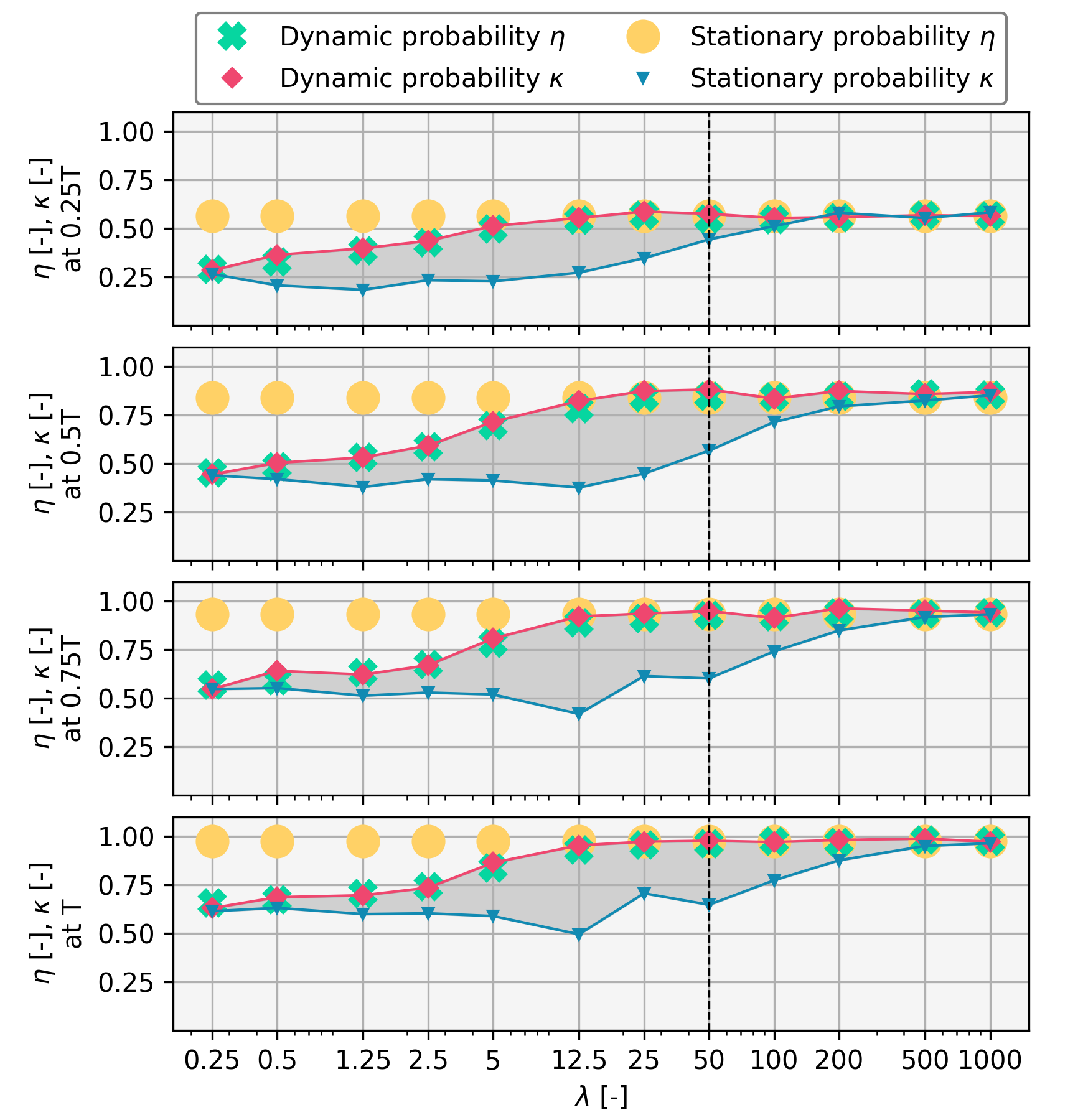}
	\caption{Survey accomplishment rate and target detection rate in relation to $\lambda$ which represents the ratio of UAV velocity and average flow field velocity. The dashed line at $\lambda=50$ represents the average estimated value for maritime search mission utilizing multirotor UAVs. The graph was generated using the synthetic cavity test case for different $\lambda$ values, by scaling the flow field. It showcases increased performance of the proposed method, compared to the baseline static probability approach, in the range of realistic values at different times of the search $T$.}
	\label{fig:cavity_analysis}
\end{figure}

As evident from Fig. \ref{fig:cavity_analysis}, the proposed method offers much better search performance than the baseline method in the realistic range of $\lambda$ values. For really high $\lambda$ values, the effects of the underlying flow field can be largely ignored and the baseline static probability search method performs the same as the proposed method. In case that the flow field has higher velocity than the search agents $\lambda<1$, both methods have similar target detection rate $\kappa$, since the flow is too chaotic and the targets essentially end up catching the agents rather than the other way around. One thing to notice here is even that $\kappa$ is similar, for the proposed method $\kappa$ corresponds to the survey accomplishment metric $\eta$, while the same cannot be said for the baseline method.  

\subsection{Realistic search case - Unije Channel search}

This test case was designed to simulate a realistic search scenario on the ocean. The exploration domain is the Unije Channel, located in the Adriatic Sea between the islands of Unije and Lošinj. The search has a 3 hour delay from the time simulated targets were placed in the ocean at known locations, matching $m_0$. The search is conducted in a total of 6 phases, with each phase representing the UAV swarm flying for the duration of one battery charge, which lasts 25 minutes. The swarm consists of 5 identical search agents, as described in Table \ref{tab:agent_charactristics}. The UAVs are deployed from the coastal region, and in between each search phase, there is a 5 minute delay simulating the UAVs' return to base and battery replacement. 

UAVs are equipped with a camera sensor that captures a rectangular area of 160 m in width, and 90 m in height, where the height direction corresponds to UAVs' heading. The probability of detecting the target within the sensor's scope during a single flyover is $\mu_b=0.75$. The matching sensing function is described with 

\begin{equation*}
	\gamma_b = -\frac{\ln(1 - \mu_b)}{9} \cdot \mu_b.
\end{equation*}

\begin{figure}[!ht]
	\centering
	\includegraphics[width=0.95\linewidth]{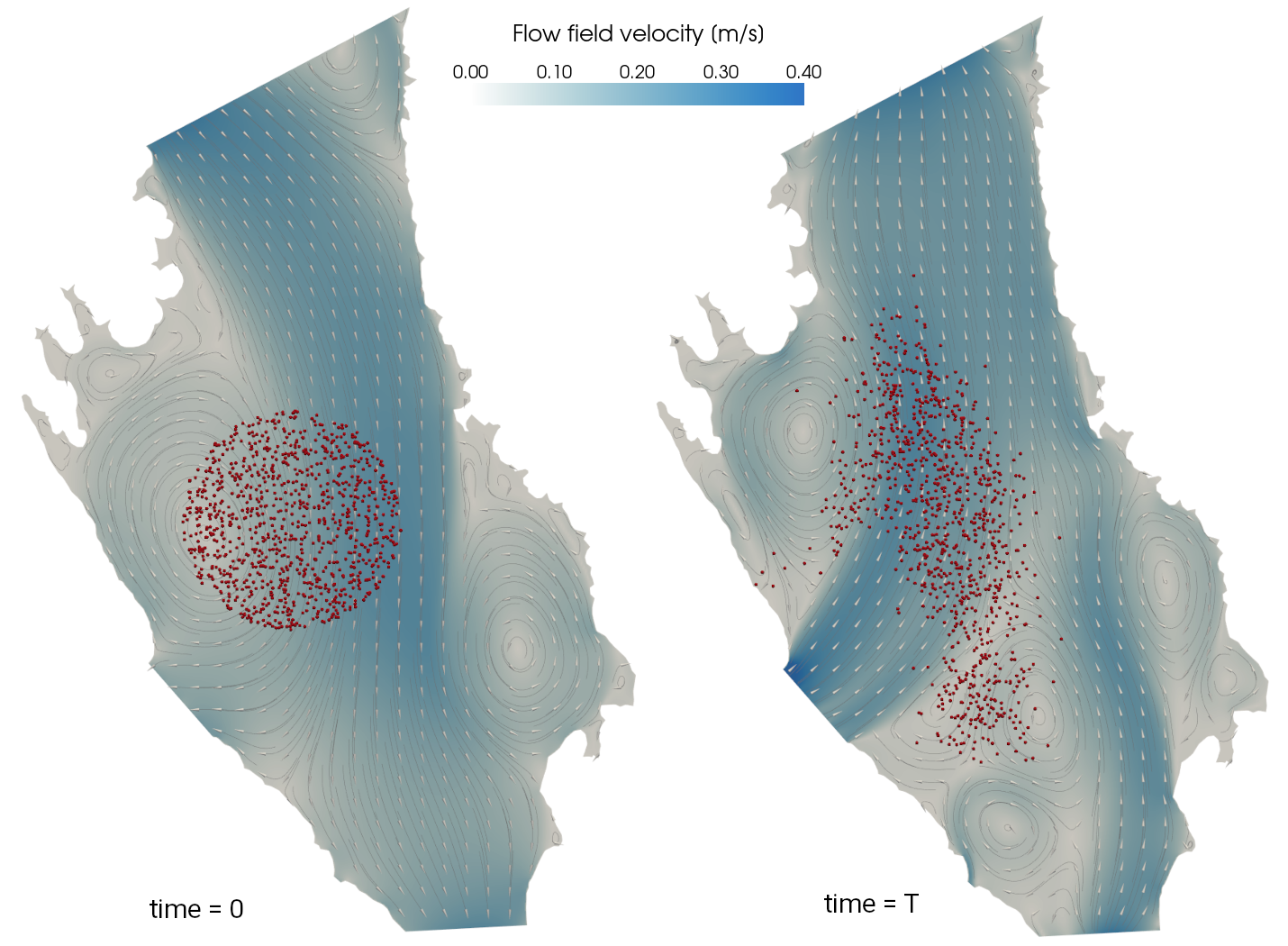}
	\caption{Visual representation of the flow field and simulated targets' positions at the start of advection $t=0$, and at the end of the search $t=T$ for the Unije Channel test case. The executed search mission is started at $t=3$ h. }
	\label{fig:unije_flow}
\end{figure}

The flow field in this scenario is transient, so it changes in respect to time, which is evident in the drift direction change of simulated targets during the simulation. The flow field for $t = 0$ and $t = T$ was generated utilizing \textit{OpenFOAM}, while the complete transient flow simulation was computed by employing the method presented in \cite{jakac2025efficient}. The velocity intensity of the generated flow field is in $[0, 0.4]$ m/s, which is equivalent to a measured real flow field surface layer velocity \cite{jakac2025efficient, cosoli2013surface}. The flow field and simulated search targets at $t=0$ and $t=T$ are shown in Fig. \ref{fig:unije_flow}.

In the scope of this search scenario we considered uncertainties that may be present in a real-world scenario. For example, the flow field may not be completely accurate, and therefore the targets drift will not precisely align with the effects of the flow filed. To simulate the uncertainties of the system, error was introduced in the advection of simulated search targets. The error for each particle was introduced by emulating the Brownian motion phenomenon defined as stochastic differential equation
\begin{equation*}
	\diff \mathbf{y}(t) = \mathbf{w}(\mathbf{y}, t)\diff t + \sigma\diff \mathbb{B},
\end{equation*}
where $\mathbb{B}$ denotes a Wiener process for standard Brownian motion with standard deviation $\sigma$.
The realistic targets' drift error was determined experimentally in \cite{jakac2025efficient}, from where we can linearly extrapolate the drift error value at 3h (half of our advection simulation time), which is roughly $e=330$ m. To replicate that effects, and achieve an error of about 330 m after 3 h of advection, we introduce error every $\Delta t = 3$ s, with a standard deviation of $\sigma = 5.4772$ m. 

\begin{figure*}[!ht]
	\centering
	\includegraphics[width=0.92\linewidth]{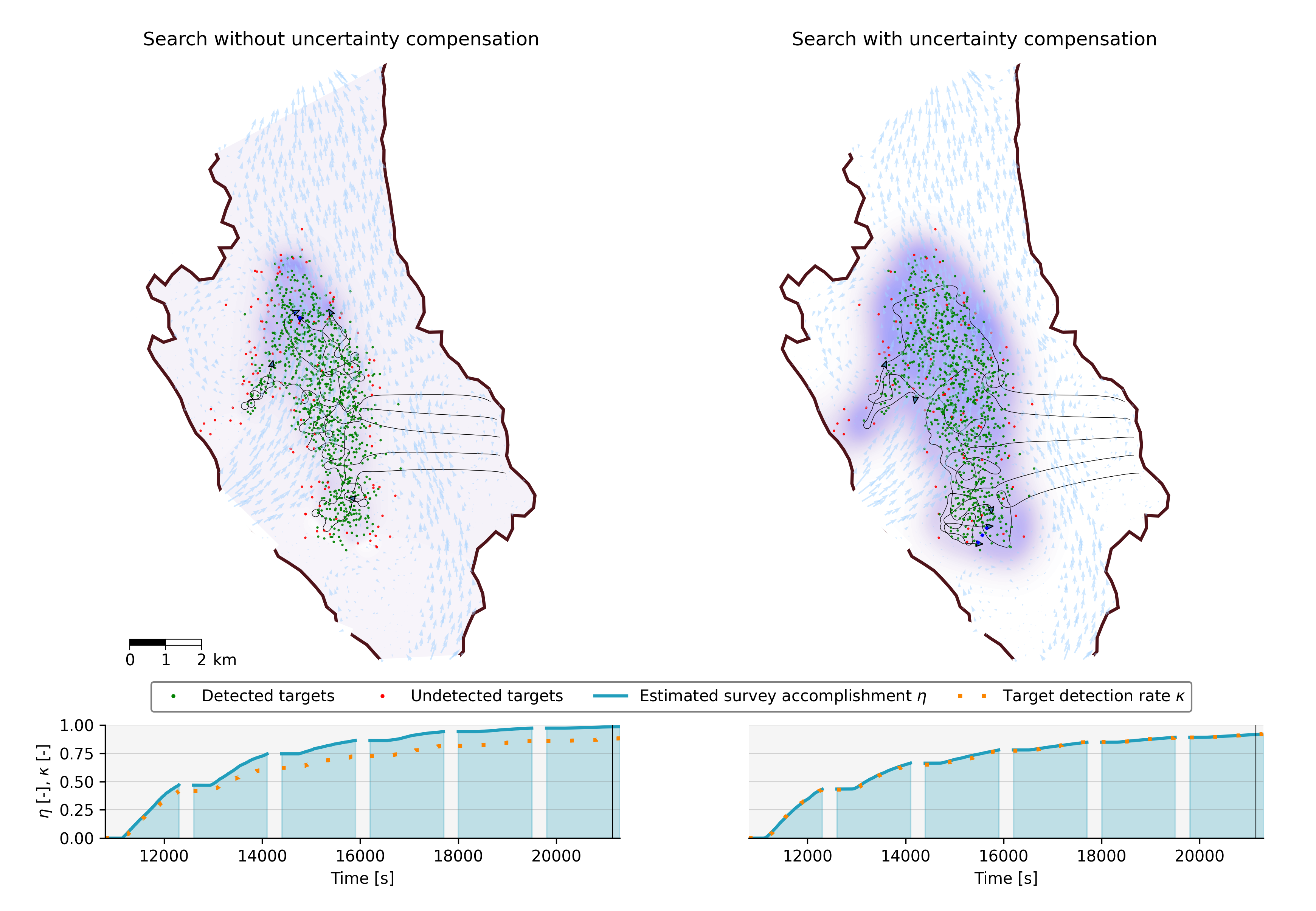}
	\caption{The Unije Channel test case search analysis including the comparison for both cases with and without uncertainty compensation. The thick black lines on the domain plots represent the coastline and the breaks in those lines represent the open water. The blue arrows in the background represent the flow field vectors, while the purple-white filled contour plot is a representation of the potential field that governs the motion of UAV search agents. The situation represents the search state at $t=21150$ s, and the UAVs' trajectories during the final search phase are represented with the thin black lines inside the search domain. The graphs below the domain plot showcase the target detection rate and survey accomplishment metrics throughout the whole duration of the search.}
	\label{fig:unije_showcase_frame}
\end{figure*}

In order to compensate for the introduced error, we calculated the diffusion coefficient by using \eqref{eq:diff_coef}, and the resulting diffusion is equal to $D=5$ $\text{m}^2/\text{s}$. The results of the search using diffusion for uncertainty compensation were compared to a case without uncertainty compensation, as shown in Fig. \ref{fig:unije_showcase_frame}. During the initial stages of the search, there is only a minor difference between the two cases, up until about 45\% of target detection rate, since in both approaches first search the area of highest target probability density. As the time goes on, the performance of the diffused case becomes better because of two reasons: firstly the case with no diffusion does not compensate for the drift error, and secondly, the target error distance increases in time. 

At the end of the search $t=T$, the method without uncertainty compensation found 883, and the method with uncertainty compensation found 922 out of 1000 total targets. That would equate to 4.4\% better search performance with uncertainty compensation. Another advantage of the of using uncertainty compensation is that the survey accomplishment metric $\eta$ is a good representation of the target detection rate $\kappa$. In that case, $\eta=0.9191$ and $\kappa=0.922$, while $\eta=0.9859$ and $\kappa=0.883$, for the case without compensation. Having an incorrect approximation of survey accomplishment could lead to wrong decision making during search mission. For example, if the mission is considered finished when $\eta=98$\%, the actual value would be 10\% lower in the uncompensated case. As a result, the next search phase, which could further increase $\eta$ and potentially locate the target, would not be executed.

\begin{figure}[!ht]
	\centering
	\includegraphics[width=1\linewidth]{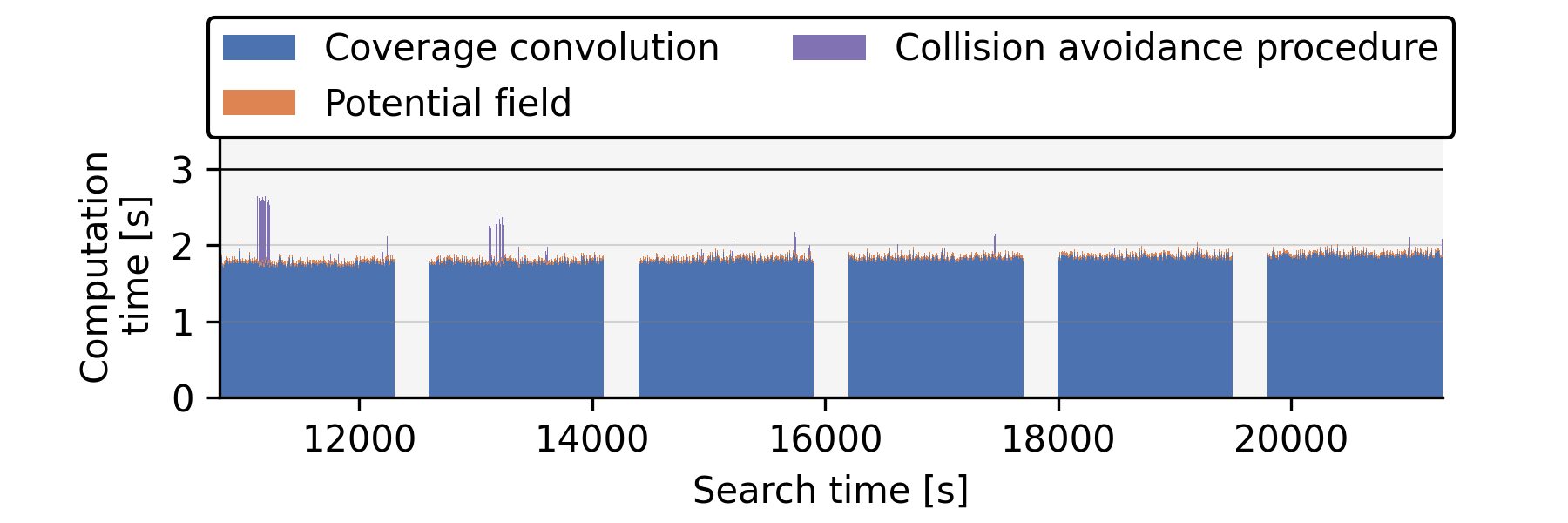}
	\caption{Computation time for the 6 simulated search phases conducted within the Unije Channel search scenario. The total computation time is divided between the potential field generation, collision avoidance procedure and the coverage convolution. The coverage convolution makes the highest contribution since it includes running the \textit{OpenFOAM} simulation for the advection of the probability field. Considering that the computation time does not exceed $\Delta t = 3$ s, we can conclude that the method is suitable for real-time UAV control.}
	\label{fig:unije_cpu_time}
\end{figure}

During the simulation, computation time was measured on same computer configuration as defined in section \ref{sec:implementation}, with the results shown in Fig. \ref{fig:unije_cpu_time}. From that we can conclude that the algorithm is suitable for real time control of UAVs since the computation time never exceeded the control time step of 3 s. The FEM mesh is composed of 18289 triangular elements and 9357 mesh nodes. The FVM mesh is comprised of 61340 hexahedral cells and 124628 points. The area represented by the search domain is equal to $95.9 \text{ km}^2$.

\section{Conclusion}

The work showcases an ergodic area exploration method with a dynamic target probability density field, aimed to be used in a maritime search scenario where the ocean flow field determines the effects of targets' motion. The method is based on the HEDAC ergodic search framework providing motion control, collision avoidance, and sensing, while \textit{OpenFOAM} scalar transport is used for advection and diffusion of the probability density field, capturing its dynamic effects. The method was compared to a baseline HEDAC approach with a static probability density field on a synthetic case, where, as expected, it demonstrated a significant performance increase. The most notable improvement was seen within the realistic ratio of UAVs' velocity and the average sea flow field velocity. It was also utilized in a simulated realistic maritime search scenario where the domain was explored by 5 UAVs equipped with sensors capturing a rectangular area imitating a camera. It featured a delayed search start time and multiple UAV flight missions with breaks considering battery replacement time. Furthermore, it demonstrated uncertainty compensation method employing the diffusion of the probability density field, which accounts for the targets' drift error, possibly caused by imperfect measurements of the flow field. The survey accomplishment of the method in both test cases was validated using simulated targets that were deployed in the flow field and drifted by its effects.

The proposed methodology provides the capability for real-time estimation of the total target detection probability and multi-agent survey control in a complex search environment. This is of significant importance for managing of resources and guiding decision-making processes in ocean search and rescue missions, where timely and accurate information is crucial for mission success.



\section*{Acknowledgments}

This publication is supported by the Croatian Science Foundation under the projects: UIP-2020-02-5090, MOBDOK-2023-1977, MOBDOK-2023-6489. It was also in part supported by the State Secretariat for Education, Research and Innovation in Switzerland for participation in the European Commission's Horizon Europe Program through the INTELLIMAN project (\url{https://intelliman-project.eu/}) and the SESTOSENSO project (\url{http://sestosenso.eu/}).

\section*{Supplementary data}
All parameters required to reproduce this study are provided in the manuscript. The data for replicating the search scenarios and video animations can be accessed via the Open Science Framework repository: \url{https://osf.io/3a6xc/}. The Python code used in this research is available upon request.


\bibliographystyle{plain}
\bibliography{bibliography}

\addtolength{\textheight}{-12cm}   

\end{document}